# What happens in Face during a facial expression? Using data mining techniques to analyze facial expression motion vectors

Mohamad Roshanzamir, Roohallah Alizadehsani, Member, IEEE, Mahdi Roshanzamir, Afshin Shoeibi, Juan M. Gorriz, Abbas Khosrave, Senior IEEE, Saeid Nahavandi, Fellow, IEEE.

*Abstract*—One of the most common problems encountered in human-computer interaction is automatic facial expression recognition. Although it is easy for human observer to recognize facial expressions, automatic recognition remains difficult for machines. One of the methods that machines can recognize facial expression is analyzing the changes in face during facial expression presentation. In this paper, optical flow algorithm was used to extract deformation or motion vectors created in the face because of facial expressions. Then, these extracted motion vectors are used to be analyzed. Their positions and directions were exploited for automatic facial expression recognition using different data mining techniques. It means that by employing motion vector features used as our data, facial expressions were recognized. Some of the most state-of-the-art classification algorithms such as C5.0, CRT, QUEST, CHAID, Deep Learning (DL), SVM and Discriminant algorithms were used to classify the extracted motion vectors. Using 10-fold cross validation, their performances were calculated. To compare their performance more precisely, the test was repeated 50 times. Meanwhile, the deformation of face was also analyzed in this research. For example, what exactly happened in each part of face when a person showed fear? Experimental results on Extended Cohen-Kanade (CK+) facial expression dataset demonstrated that the best methods were DL, SVM and C5.0, with the accuracy of 95.3%, 92.8% and 90.2% respectively.

M. Roshanzamir is with the Department of Computer Engineering, Faculty of Engineering, Fasa University, 74617-81189, Fasa, Iran (email: roshanzamir@fasau.ac.ir).

R. Alizadehsani and A. Khosravi are with the Institutefor Intelligent Systems Research and Innovation (IISRI), Deakin University,Victoria, Australia (emails: r.alizadehsani@deakin.edu.au, abbas.khosravi@deakin.edu.au).

M. Roshanzamir is Department of Electrical and Computer Engineering, University of Tabriz, Tabriz, Iran (email: roshanzamir@tabrizu.ac.ir)

A. Shoeibi is with the Faculty of Electrical Engineering, Biomedical Data Acquisition Lab (BDAL), K. N. Toosi University of Technology, Tehran, Iran, Iran. (e-mail: afshin.shoeibi@gmail.com)

Juan M. Gorriz is with the Department of Signal Theory, Networking and Communications, Universidad de Granada, Spain. Also with the Department of Psychiatry. University of Cambridge, UK (email: gorriz@ugr.es).

S. Nahavandi. is with the Institute for Intelligent Systems Research and Innovation (IISRI), Deakin University,Victoria, Australia. Also with the Harvard Paulson School of Engineering and Applied Sciences, Harvard University, Allston, MA 02134 USA (email: saeid.nahavandi@deakin.edu.au).

## I. INTRODUCTION

FACIAL expressions play a very important role in human interaction. This is one of the fastest and most efficient ways in communication. So, the effect of this type of communication is more than others. Mehrabian [1] indicated that emotions appear 7% in spoken words and 38% in vocal utterances, whereas facial expressions represent 55% of it. The same fact has also been mentioned in [2] as the nonverbal information prevails over words in human communication. Nowadays, humans can not only communicate with each other, but also they communicate with machines. It is not known if machines will be able to recognize emotions like those of human. Automatic facial expression recognition helps to do that to a great extent. Research fields such as image processing, machine vision, machine learning and data mining can be considered as the basis for this research. A wide range of applications can use the results of researches in this field. Human-computer interaction is probably the most important application in automatic facial expression recognition. The universal use of computational systems requires improved human-computer interaction. However, other fields such as date-driven animation, psychology researches, medicine, security, education and distance learning, customer satisfaction and video conferences can use the results of these researches. They can even be used in the (re)production of artificial emotions.

To detect facial expressions and make them available for usage in various applications mentioned above, automatic recognition of facial expression is needed. While facial expression recognition is extremely easy for human, a powerful system that can do this like human has not been built yet and it encounters many restrictions, putting it in faraway distance from the ideal model [3]. The ideal automatic facial expression recognition system must be completely automatic, person independent and robust to any environmental condition [4]. To do that, a three-stage process must be done. They are face detection, facial feature extraction, and facial expression classification [2]. As these three stages are wide, it is better to conduct research on them separately. In this paper, face detection was not addressed. We assumed that face images were already available in suitable conditions. This research only focused on facial feature extraction and facial expression classification.

The novelties of this paper are as follows:
1. This paper focused on the classification of human emotions from motion vectors extracted from image sequence of facial expressions; the third stage. Different types of data mining techniques were used and compared for this purpose. These techniques classified motion vectors extracted from a video sequence into a basic emotion.
2. The motion vector dataset used for classification is new and generated for this research.
3. Meanwhile, different changes in different parts of face were analyzed to address what exactly happened in face when a person shows an emotion. Accordingly, some changes happen in the face because of facial expression were introduced in this research for the first time.

After a presentation of the related work in Section 2, we give an overview of dataset used in our research in Section 3. Sections 4 focuses on face segmentation and motion vector extraction and finally, concentrates on data mining techniques for the classification of emotions. Section 5 describes implementation and the results. Finally, section 6 closes the paper with a summary and an outlook.

## II. RELATED WORKS

Facial expression recognition is a kind of pattern recognition problem. During the last decade, a lot of approaches for facial expression recognition have been proposed. Some of them are classical approaches. For example, in [5] and [6], support vector machine (SVM) and other features are used for facial expression recognition. In order to enhance the performance of facial expression recognition, Xiaoming et al. [7] used Gabor feature and Histogram of Oriented Gradients (HOG) with SVM classifier. They first used Gabor Wavelet filter for feature extraction. After that binary encoding (BC) and HOG algorithms were used for dimensionality reduction. Finally, support vector machine classifier was used for expression classification. In [8], a framework, which extracts features from active facial patches was used and then the expressions were classified by SVM. Facial-based features were used by Kirana et al. [9] to detect face and recognize emotion. They applied rectangular feature and cascading AdaBoost algorithm which are the main concept of the Viola-Jones Algorithm in those both of process. Fuzzy inference system is a common strategy for automatic facial expression recognition [10-12]. Finally, some of the research proposed novel classification algorithm in this field [13].

Different feature extraction techniques were combined with various classification algorithms in [14]. The main purpose of these combinations was to find the best combination that can be used for emotion recognition. Besides mentioned studies, other approaches such as PCA [15], Haar feature selection technique (HFST) [16], K-means clustering [17], feature distribution entropy [18], distance vectors [19], stochastic neighbor embedding [20], local binary patterns (LBP) [21, 22], gradient-based ternary texture patterns [23, 24], Gaussian curvature [25], spectral regression [26], Dynamic Bayesian network [27], Hidden Markov Models [28-30], conditional random field model [31] were used for facial expression recognition.

Recently, using artificial neural networks and deep learning based methods has been increased for facial expression recognition. Qin et al. [32] combined Gabor wavelet transform and convolutional neural network to increase the accuracy of traditional expression recognition methods. In [33], three deep networks were designed to recognize emotions by synchronizing speech signals and image sequences. An eleven-layered Convolutional Neural Network with Visual Attention was proposed for facial expression recognition [34]. In this approach, three components were integrated into a single network which can be trained in an end-to-end scheme. These three components were used to extract local convolutional features of faces, determine the regions of interest and finally infer the emotional label.

In order to overcome the over-fitting problem in deep learning, Binarized Auto-encoders (BAEs) and Stacked Binarized Auto-encoders (Stacked BAEs) were used to learn a kind of domain knowledge from a large-scale unlabeled facial dataset [35]. So, the performance of Binarized Neural Networks (BNNs) can be improved. An end-to-end network with an attention model for facial expression recognition was proposed by Fernandez et al. [36]. It focused attention in the human face and used a Gaussian space representation for expression recognition. For learning multi-level facial expression features and temporal dynamics of facial expressions in a joint way, a network termed Spatio-Temporal Convolutional features with Nested LSTM (STC-NLSTM) was proposed in [37]. In this study, 3DCNN was used to extract spatio-temporal convolutional features from the image sequences that represent facial expressions, and the dynamics of expressions are modeled by Nested LSTM.

In [38], a semi-supervised deep belief network (DBN) approach was proposed to determine the facial expressions. Meanwhile, a gravitational search algorithm (GSA) is applied to optimize some parameters in the DBN network to achieve an accurate classification of the facial expressions. A deep learning approach based on attentional convolutional network that was able to focus on important parts of the face and achieves significant improvement was proposed by Minaee et al. [39]. They used a visualization technique that was able to find important facial regions to detect different emotions based on the classifier's output. A modified version of the Cat Swarm Optimization (CSO) algorithm was applied to an approach for human facial expression recognition. Deep features present in the face image were extracted using Deep Convolution Neural Network (DCNN). Meanwhile, CSO was used to select optimal features in order to distinguish the facial expression of a person. Different surveys and reviews for facial expression recognition methods and deep facial expression recognition are presented in [40-45].

In [46], a feature-based approach for facial expressions recognition is introduced. It presents a fully automatic solution to identify human facial expressions. Facial features components were automatically detected and segmented. Then some points are defined as points of interest surrounding the segmented features. Some distances between these points of interest were computed. By relying on data mining robustness, these distances were used to generate a set of relevant prediction rules able to classify facial expressions effectively. Tripathi M et al. [47] used the data mining statistical approach using information gain and gini index in order to recognize the correct emotion. In [48], a new facial-expression analysis system has been introduced to manage facial-expression intensity variation and reduce the doubt and confusion between facial-expression classes. It segmented facial feature contours using Vector Field Convolution technique. By relying on the detected contours, facial feature points which went with facial-expression deformations were extracted. A set of distances among the detected points were used to define prediction rules through data mining techniques.

Although different methods such as hidden Markov model [49-51], supported vector machine [52, 53] and artificial neural network [54, 55] are widely used for automatic facial expression recognition, there are not many researches analyzing changes in face happened during facial expression using data mining techniques. In this research, the results of this analyzing is used but not limited to automatic facial expression recognition.

## III. DATASET

The first issue in automatic facial expression is finding a test dataset. It is essential to have a good test data to check new approaches and compare them with other methods and the previous researches. In this research, an image sequence of facial expression must be available. So, we selected Extended Cohen-Kanade (CK+) facial expression dataset [56] as our benchmark. It is one of the most comprehensive datasets captured and prepared under laboratory conditions. It contains 593 frontal-view image sequences from 123 posers ranging from 18 to 50

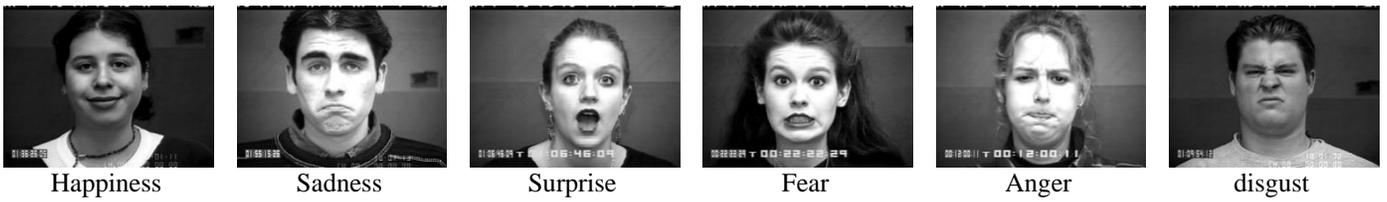
Figure 1. Six basic emotions

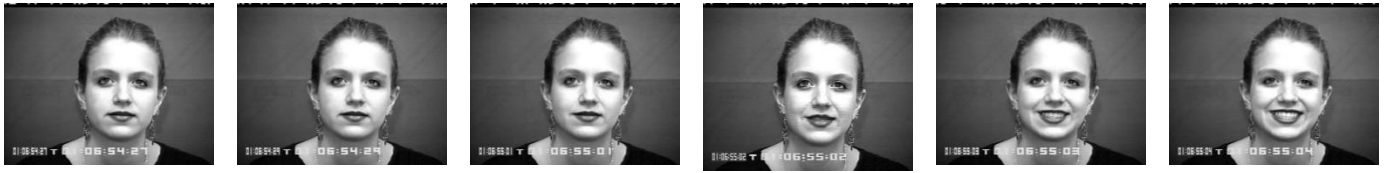
Figure 2. An image sequence that shows happiness

years of age with a variety of genders and heritage such as African-American, Asian, Latino and Americans. Image dimension is 640 × 480 or 640 × 490 pixels. The image sequences in this dataset start from the neutral state and end in peak of one of six basic emotions [57]. Basic emotions are shown in Figure 1. They include happiness, sadness, surprise, fear, anger and disgust. All six expressions sequences were not available for some subjects. Figure 2 shows an example of image sequence in CK+ dataset.

IV. DATA MINING TECHNIQUES FOR AUTOMATIC FACIAL EXPRESSION RECOGNITION

Data mining is defined as the process of automatic exploration and extraction of the knowledge from the data [58]. It is a topic of interest that involves *learning* in a practical sense. The word *learning* means that these techniques learn from the changes appearing in the data in a way that improves their performance in the future. Thus, *learning* is tied to performance enhancement [59]. Based on this learning process, the learning techniques can be employed to map data into decision model in order to produce the predicting output from the new data. This decision model is called classifier [60]. Now a days, these techniques used a lot in different fields [61-63].

In this section, at first, the method used for data collection was introduced in subsections A and B. Then, data cleaning and classification were explained as shown in subsection C.

*A. Motion vector extraction*

In this stage, data collection was done. The collected data were motion vectors. A facial expression results in some temporary shift of facial features because of facial muscle movements. These deformations are short in time and take only a few seconds. Motion vectors that show facial deformation were extracted from image sequence of facial expression. There are some algorithms for this purpose such as optical flow and different tracking algorithms. In this research, optical flow algorithm [64] was used to extract motion vectors. It is based on tracking points across multiple images. In this research, at most, 8-image sequence was used for each test. We used image sequences of faces in a frontal view displaying various facial expressions of emotion. The pre-processing of the images was performed by converting them into gray scale images and then segmenting the face in a rectangular bounding. Extra parts of images were cut. It resizes image dimensions to about 280 × 330 pixels. We also assume that the input to our system consisted of facial expression sequences that always started with a neutral facial expression and ended with the apex of a facial expression. The faces were without hair and glasses and no rigid head movement could be acceptable for the method to work properly. An example of image modification is illustrated in Figure 3(a) and Figure 3(b). They show the image sequence of disgust and happiness, respectively.

Optical flow algorithm has some weaknesses. For example, luminance must not change while the image sequence is created. Otherwise, this algorithm is not able to extract motion vector correctly. However, as the changes in face because of facial expression happens in a very short time, commonly luminance change was not happen. So, this weakness of optical flow is not a critical problem in this algorithm. Additionally, in CK+ dataset, this condition is observed in all cases. However, there are a few cases the subject has rigid head movement while they show an expression. This disturbs the extracted motion vectors and will mislead the classification algorithms. So, we eliminated these types of extracted motion vectors. Different modifications have been done on optical flow algorithm. To minimize the effects of luminance variation and inaccuracies in facial point tracking, Gautama-VanHulle optical flow method [65] has been used. It was claimed that this method is less sensitive to luminance variation and has very good efficiency in motion vector extraction. Two examples of using optical flow algorithm on image sequences of facial expression have been presented in Figure 3 (c) and Figure 3 (d). These two figures show disgust and happiness motion vectors, respectively.

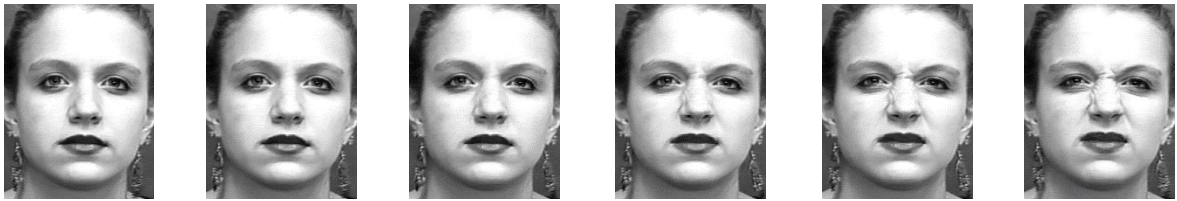
a

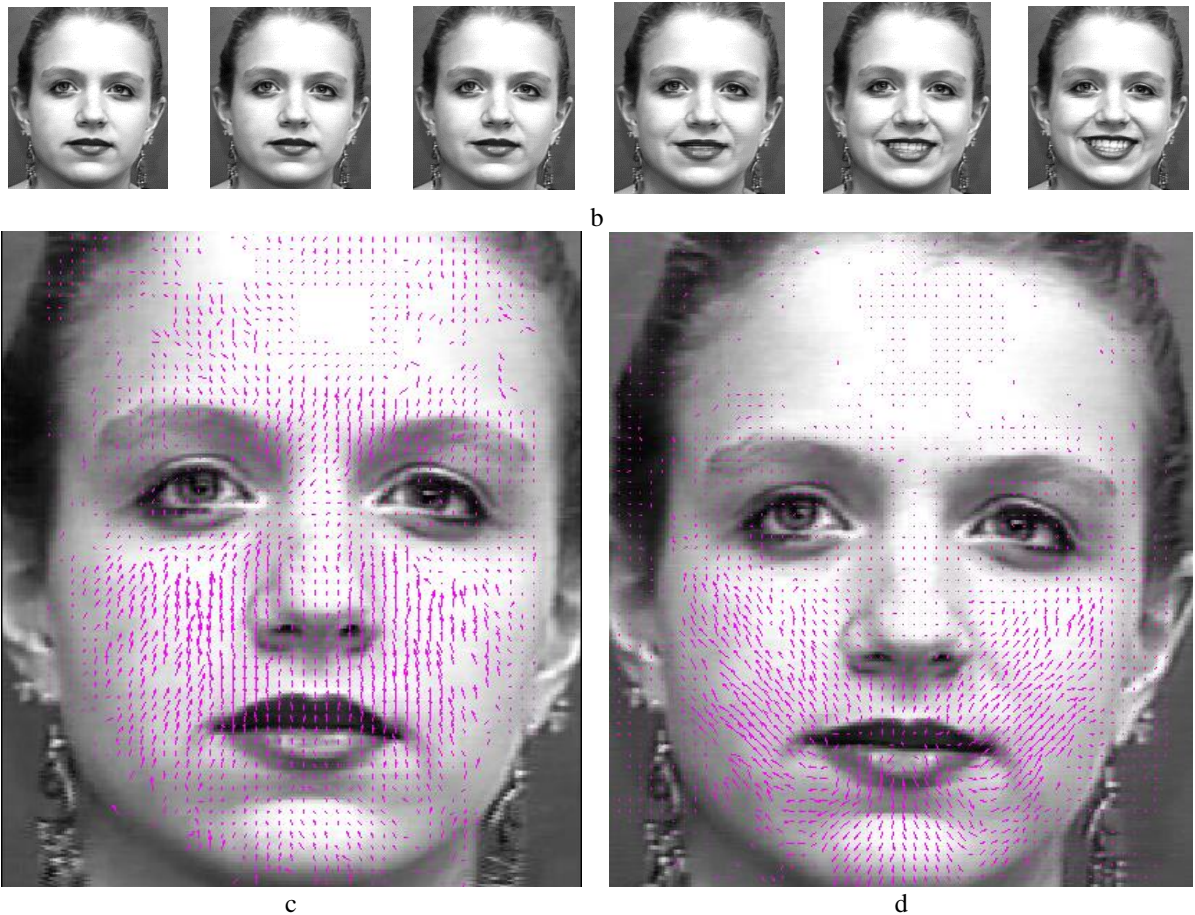

Figure 3. Motion vectors extracted using optical flow algorithm. (a) shows the image sequence of disgust (b) shows the image sequence of happiness. (c) Disgust motion vectors extracted from disgust image sequence and (d) Happiness motion vectors extracted from happiness image sequence.

*B. Face segmentation*

To classify extracted motion vectors into six basic emotions, the face was divided into six parts as shown in Figure 4(a). A similar segmentation has been done in [13]. At first, the position of pupils and mouth must be located. Their initial location can be detected either manually or automatically in the first frame of the input face image sequence. In this research, their initial locations were selected manually. However, it is possible to detect them automatically with high accuracy [66-69]. As it is clear in Figure 4(b), axis number 1 connects two pupils. Axis number 3 is perpendicular to axis number 1 and divides it into two equal parts. Axis number 2 shows the mouth position. It is not necessary to specify axes location precisely. Approximation of axis locations is enough.

Figure 5 and Figure 6 show why such a kind of face segmentation is done. Figure 5 shows Bassile facial expression deformation [70] while Figure 6 displays other types of deformations extracted experimentally from facial expression image sequences in CK+ dataset. It is clear from these figures that the most important changes happen on the top of eyes, around the eyebrows and mouth. So these areas are divided into smaller sections as shown in Figure 7(a). Dividing these areas into smaller sections gives the chance to analyze the deformations in these sections more precisely. In Figure 7(b), nine vectors showing different directions and segmentations can be seen. As face is symmetric, the number and size of segments in $X_1$ and $X_4$, $X_2$ and $X_5$ and, $X_3$ and $X_6$ directions are the same. These segmentations can be different in width, length and number. In each segment, the ratio of vector numbers and their average length in x and y directions are extracted and used for mining. So, for each image sequence, 3n features are extracted, where n is the number of small segments. For example, in Figure 7(a), the number of small segments is 120. So 360 features are extracted and used for mining. By using this 360 feature, the system should analyze the face deformation and detect facial expressions.

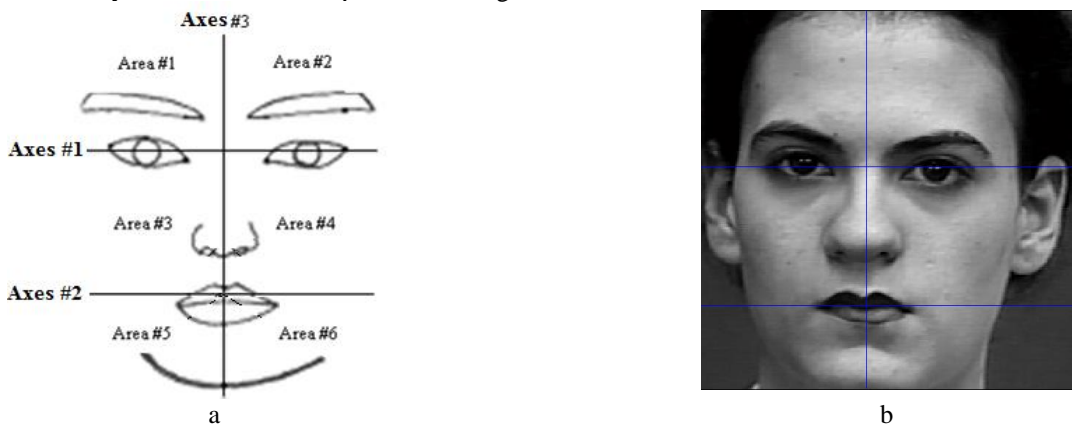

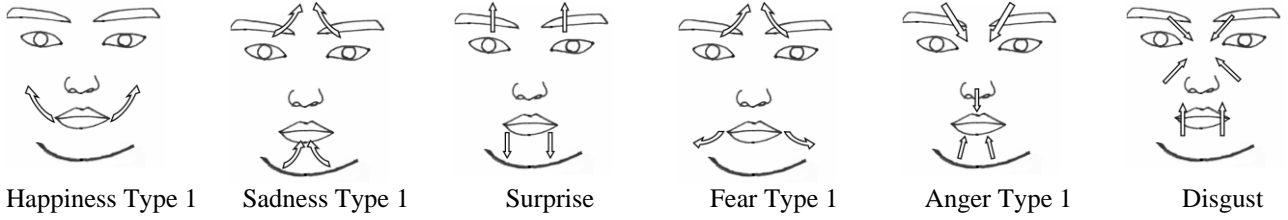

Figure 4. Face segmentation into 6 areas in a a)scheme and b) real face image

| Happiness Type 1 | Sadness Type 1 | Surprise | Fear Type 1 | Anger Type 1 | Disgust |

Figure 5. Bassili description of face deformation in each emotion

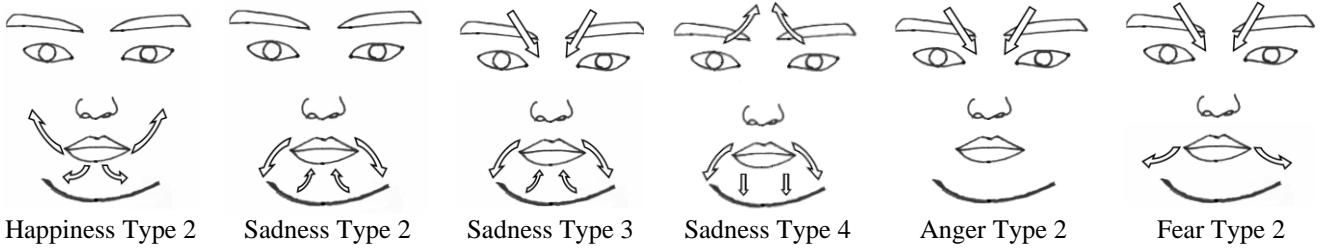

| Happiness Type 2 | Sadness Type 2 | Sadness Type 3 | Sadness Type 4 | Anger Type 2 | Fear Type 2 |

Figure 6. Other types of face deformation in different emotions

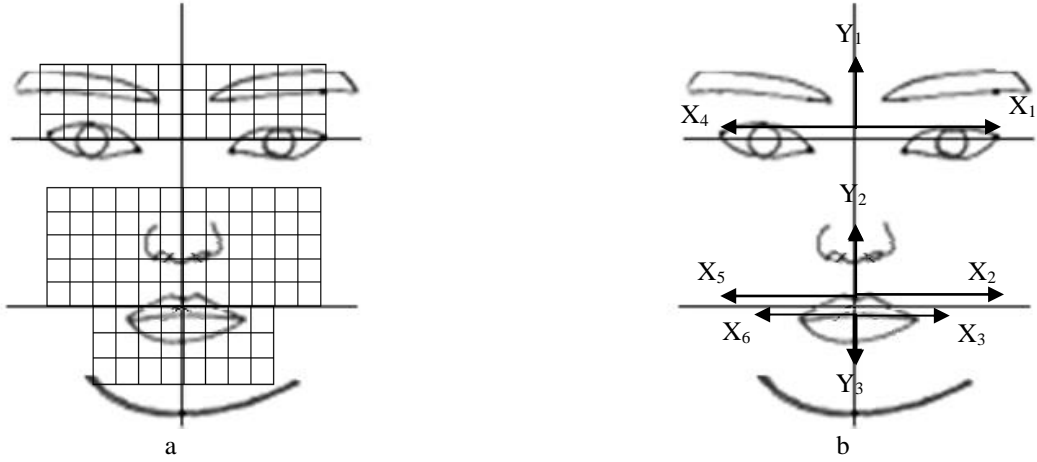

Figure 7. a) Face areas having the most important deformation during facial expression are divided into smaller sections. b) Nine different directions in which segmentations are done.

*C. Facial expression classification*

In this stage, extracted data must be analyzed. At first, data had to be cleaned and outlier, extreme and missing values must be handled. Outliers were replaced with the nearest value that would not be considered as an outlier. For example, if an outlier was defined to be anything above or below three standard deviations, then all of them would be replaced with the highest or lowest value within this range. Extreme values would be discarded and missing values would be replaced by zero. These missing values were created in sections wherein there was no motion vector. So, zero was found to be the best value for substitution of them. Then, 10-fold cross validation was applied 50 times on these data. In this research, C5.0 [71], CRT [72], QUEST [73], CHAID [74], Neural Networks [75], Deep Learning (DL) [75, 76], SVM [77] and Discriminant algorithms [78] were used to classify the extracted features. These classification algorithms were used to extract knowledge from a dataset consisting of motion vector features. These motion vectors formed feature vectors. These features calculated according to equations 1, 2 and 3.

$$P_{ij} = \frac{N_{ij}}{N} \qquad 1$$

$$LX_{ij} = \frac{1}{N_{ij}} \sum_{k=1}^{N_{ij}} lx_{ij}^k \qquad 2$$

$$LY_{ij} = \frac{1}{N_{ij}} \sum_{k=1}^{N_{ij}} ly_{ij}^k \qquad 3$$

where

- $N$ is the number of extracted motion vectors.
- $N_{ij}$ is the number of extracted motion vectors in section *ij*.
- $P_{ij}$ is the ratio of motion vector numbers in section *ij*.
- $lx_{ij}^k$ is the length of motion vector $k$ in direction $x$ in section *ij*.
- $ly_{ij}^k$ is the length of motion vector $k$ in direction $y$ in section *ij*.
- $LX_{ij}$ is the mean of motion vectors' length in direction $x$ in section *ij*.
- $LY_{ij}$ is the mean of motion vectors' length in direction $y$ in section *ij*.

As illustrated in Figure 8, each feature vector was composed of the values of features calculated by equations 1, 2 and 3. The ratio of vector number and the mean length of motion vectors in each created section were used as characteristic features.

| $P_{11}$ | $LX_{11}$ | $LY_{11}$ | $P_{12}$ | $LX_{12}$ | $LY_{12}$ | $\cdots$ | $P_{NM}$ | $LX_{NM}$ | $LY_{NM}$ |

Figure 8. An example of feature vector used for facial expression classification.

The location and size of deformation are often different from one image sequence to another. They even differ from one occurrence to another one in a person. However, these differences are not a lot in a specific facial expression. Pseudo-code of the proposed method can be seen in **Error! Not a valid bookmark self-reference.**.

Algorithm 1. Pseudo-code of the proposed method

1. Prepare image sequence    % Onset: natural state;
                             % Apex: one of six basic emotions.
2. Apply preprocessing on image sequence
   2.1. Convert images into grayscale
   2.2. Cut images so that only the face is shown in them.
3. Extract motion vectors in image sequence using optical flow algorithm.
4. Divide images into six areas.
5. Segment each area as shown in Figure 7(a)
6. Extract data
   6.1. Calculate the percentage of the number of vectors in each segment
   6.2. Calculate the mean of vector length in x and y directions in each segment
7. Eliminate outlier and missing data from extracted motion vectors.
8. For 50 times,
   8.1. Apply 10-fold cross-validation
   8.2. Estimate the performance of classifier
9. Calculate the overall performance of classifier

## V. EXPERIMENTAL RESULTS

In order to evaluate and compare the performance of different algorithms mentioned in Section 4, motion vectors were extracted from 475 image sequences of CK+ dataset. They represented one of the facial expressions shown in Figure 5 and Figure 6. The location of motion vectors and their length in x and y directions were saved and used for extraction of feature vectors used in different classification algorithms. For each image sequence, between 18 and 631 features were used and for each facial expression, 50 times 10-fold cross validation were applied. Different algorithms were tested on 25 situations. These situations had different width, height and number of segments in different directions shown in Figure 7(b). These situations have different numbers of segments and sizes as shown in Table 1. They were selected empirically so that motion vectors in each segment would have the same direction according to Figure 4 and Figure 5. To determine how many segments were more suitable and what size is better for them, these different situations were compared. The results are shown in Table 2.

Table 1. Features of different situations. Segments width and height are in pixel

| | | Segments | | Number of Segments in different directions | | | | | | Total number of features |
|---|---|---|---|---|---|---|---|---|---|---|
| | | Width | Height | $X_1$ & $X_4$ | $Y_1$ | $X_2$ & $X_5$ | $Y_2$ | $X_3$ & $X_6$ | $Y_3$ | |
| Situation | 1 | 5 | 5 | 3 | 3 | 4 | 3 | 2 | 3 | 162 |
| | 2 | 5 | 5 | 8 | 5 | 8 | 5 | 5 | 5 | 630 |
| | 3 | 5 | 15 | 2 | 1 | 3 | 1 | 2 | 1 | 42 |
| | 4 | 5 | 15 | 8 | 2 | 8 | 2 | 5 | 2 | 252 |
| | 5 | 5 | 20 | 5 | 2 | 5 | 2 | 2 | 1 | 132 |
| | 6 | 10 | 10 | 5 | 4 | 5 | 4 | 5 | 3 | 330 |
| | 7 | 10 | 10 | 5 | 5 | 5 | 5 | 5 | 5 | 450 |
| | 8 | 10 | 10 | 5 | 6 | 5 | 6 | 5 | 4 | 480 |
| | 9 | 10 | 15 | 5 | 3 | 5 | 3 | 3 | 2 | 216 |
| | 10 | 15 | 15 | 4 | 3 | 4 | 3 | 3 | 2 | 168 |
| | 11 | 15 | 15 | 5 | 5 | 5 | 5 | 5 | 5 | 450 |
| | 12 | 15 | 20 | 3 | 2 | 3 | 2 | 2 | 1 | 84 |
| | 13 | 20 | 20 | 5 | 5 | 5 | 5 | 5 | 5 | 450 |
| | 14 | 20 | 25 | 2 | 1 | 2 | 1 | 1 | 1 | 30 |
| | 15 | 25 | 25 | 2 | 2 | 2 | 2 | 2 | 2 | 72 |
| | 16 | 25 | 25 | 4 | 4 | 4 | 4 | 4 | 4 | 288 |
| | 17 | 25 | 30 | 1 | 1 | 1 | 1 | 1 | 1 | 18 |
| | 18 | 30 | 30 | 2 | 2 | 2 | 2 | 1 | 1 | 54 |
| | 19 | 30 | 30 | 2 | 2 | 2 | 2 | 2 | 2 | 72 |
| | 20 | 30 | 30 | 3 | 3 | 3 | 3 | 2 | 2 | 132 |
| | 21 | 30 | 30 | 4 | 4 | 4 | 4 | 4 | 4 | 288 |
| | 22 | 35 | 35 | 3 | 3 | 3 | 3 | 3 | 3 | 162 |
| | 23 | 40 | 40 | 3 | 3 | 3 | 3 | 2 | 2 | 132 |
| | 24 | 40 | 40 | 3 | 3 | 3 | 3 | 3 | 3 | 162 |
| | 25 | 50 | 50 | 2 | 2 | 2 | 2 | 1 | 1 | 54 |

Table 2. Accuracy (%) of different algorithms in different situations. The best situation in each algorithm is bolded.

| | | Algorithms | | | | | | | |
|---|---|---|---|---|---|---|---|---|---|
| | | DL | CRT | QUEST | CHAID | SVM | C5.0 | Discriminant | *Overall AVG* |
| Situation | 1 | 81.4 ± 16.6 | 52.9 ± 13.5 | 52.9 ± 4.8 | 58.2 ± 6.6 | 75.5 ± 21.6 | 88.1 ± 21.2 | 63.9 ± 11.8 | *67.5 ± 15.0* |
| | 2 | 30.3 ± 22.3 | 59.9 ± 11.4 | 63.0 ± 18.7 | 65.9 ± 8.5 | 78.8 ± 17.1 | 76.9 ± 7.3 | 13.9 ± 20.4 | *55.5 ± 16.1* |
| | 3 | 75.7 ± 8.9 | 52.1 ± 19.9 | 51.5 ± 15.9 | 53.3 ± 15.1 | 77.4 ± 5.6 | 89.1 ± 16.0 | 65.2 ± 12.1 | *66.3 ± 14.1* |
| | 4 | 86.7 ± 6.2 | 58.2 ± 20.5 | 63.1 ± 7.3 | 73.2 ± 17.0 | 80.3 ± 21.5 | 81.0 ± 15.2 | 64.6 ± 10.1 | *72.4 ± 15.1* |
| | 5 | 83.3 ± 12.2 | 61.9 ± 5.5 | 58.9 ± 17.5 | 69.0 ± 16.0 | 77.7 ± 9.7 | 87.2 ± 18.0 | 71.3 ± 11.3 | *72.8 ± 13.6* |
| | 6 | 87.5 ± 7.8 | 64.1 ± 8.6 | 71.6 ± 20.7 | 77.4 ± 16.9 | 83.7 ± 6.5 | 86.6 ± 6.7 | 61.9 ± 8.7 | *76.1 ± 12.0* |
| | 7 | 90.4 ± 17.9 | 69.5 ± 8.3 | 71.2 ± 9.3 | 74.9 ± 8.9 | 82.4 ± 5.3 | 69.5 ± 7.5 | 58.3 ± 8.3 | *73.7 ± 10.1* |
| | 8 | 89.2 ± 10.9 | 67.6 ± 10.3 | 72.5 ± 22.6 | 75.1 ± 18.8 | 84.4 ± 5.7 | 83.4 ± 10.3 | 57.6 ± 22.2 | *75.7 ± 15.7* |
| | 9 | 88.8 ± 8.0 | 65.3 ± 17.0 | 72.7 ± 10.8 | 76.9 ± 5.9 | 87.9 ± 5.2 | 73.7 ± 11.2 | 63.3 ± 6.9 | *75.5 ± 10.0* |
| | 10 | 91.5 ± 17.6 | 67.9 ± 9.2 | 76.0 ± 18.3 | 75.7 ± 9.4 | 87.5 ± 12.6 | 78.6 ± 5.6 | 63.6 ± 24.6 | *77.2 ± 15.2* |
| | 11 | 91.4 ± 8.5 | 67.9 ± 6.0 | 69.3 ± 17.7 | **79.7 ± 4.9** | 87.3 ± 6.1 | 65.5 ± 5.8 | 64.3 ± 12.5 | *75.0 ± 9.8* |
| | 12 | 86.4 ± 9.1 | 61.7 ± 15.2 | 67.5 ± 13.7 | 68.2 ± 19.2 | 76.2 ± 15.1 | 84.7 ± 7.6 | 84.0 ± 9.9 | *75.5 ± 13.4* |
| | 13 | 89.7 ± 12.3 | 67.3 ± 8.6 | 74.3 ± 10.4 | 76.9 ± 16.4 | 85.4 ± 19.3 | 64.7 ± 24.4 | 61.6 ± 9.3 | *74.3 ± 15.4* |
| | 14 | 84.4 ± 9.3 | 56.6 ± 13.7 | 68.9 ± 17.6 | 65.8 ± 12.8 | 81.3 ± 7.6 | 88.3 ± 10.0 | 76.9 ± 12.6 | *74.6 ± 12.3* |

| | 15 | 90.0 ± 12.5 | 64.5 ± 15.0 | 70.5 ± 12.9 | 72.4 ± 7.7 | 85.4 ± 23.2 | 84.3 ± 9.0 | 82.2 ± 24.5 | 78.5 ± 16.2 |
|---|---|---|---|---|---|---|---|---|---|
| | 16 | 91.3 ± 12.4 | 69.7 ± 8.6 | **76.7 ± 20.8** | 72.1 ± 17.9 | 86.6 ± 7.6 | 72.1 ± 23.1 | 67.8 ± 5.9 | *76.6 ± 15.2* |
| | 17 | 81.2 ± 21.8 | 61.9 ± 15.4 | 61.6 ± 17.0 | 62.9 ± 16.3 | 88.3 ± 11.6 | 86.4 ± 24.2 | 76.4 ± 17.3 | *74.1 ± 18.1* |
| | 18 | 89.1 ± 15.9 | 69.1 ± 4.7 | 72.3 ± 6.3 | 71.2 ± 10.5 | 82.4 ± 17.3 | 87.6 ± 14.5 | 81.7 ± 24.0 | *79.1 ± 14.7* |
| | 19 | 92.8 ± 22.8 | **74.4 ± 4.7** | 71.6 ± 22.6 | 77.3 ± 23.4 | **92.8 ± 17.4** | **90.2 ± 17.3** | 81.2 ± 13.1 | ***82.9 ± 18.4*** |
| | 20 | 88.6 ± 17.1 | 69.8 ± 20.9 | 70.6 ± 22.0 | 75.7 ± 20.6 | 85.9 ± 15.7 | 78.5 ± 13.6 | 71.4 ± 14.0 | *77.2 ± 18.0* |
| | 21 | **95.3 ± 18.7** | 67.9 ± 5.8 | 70.6 ± 7.5 | 74.6 ± 20.3 | 83.2 ± 15.0 | 73.4 ± 16.2 | 67.5 ± 12.1 | *76.1 ± 14.6* |
| | 22 | 88.3 ± 20.4 | 64.3 ± 14.3 | 74.9 ± 6.7 | 76.7 ± 17.6 | 82.9 ± 13.3 | 72.1 ± 8.5 | 67.9 ± 18.9 | *75.3 ± 15.0* |
| | 23 | 92.7 ± 5.4 | 68.9 ± 23.0 | 71.8 ± 19.4 | 72.7 ± 21.4 | 87.8 ± 10.5 | 76.6 ± 9.4 | 77.2 ± 11.5 | *78.2 ± 15.7* |
| | 24 | 91.4 ± 18.1 | 68.9 ± 20.2 | 70.1 ± 8.7 | 78.9 ± 20.4 | 86.7 ± 18.3 | 70.6 ± 24.1 | 69.1 ± 22.9 | *76.5 ± 19.5* |
| | 25 | 87.8 ± 20.9 | 66.9 ± 11.6 | 70.4 ± 12.0 | 74.0 ± 22.7 | 81.0 ± 7.0 | 89.1 ± 6.5 | **83.8 ± 19.5** | *79.0 ± 15.6* |
| Overall AVG | | **85.8 ± 15.1** | *64.8 ± 13.6* | *68.6 ± 15.5* | *71.9 ± 16.0* | *83.6 ± 13.9* | *79.9 ± 14.7* | *67.9 ± 15.7* | |

It is clear from Table 2 that DL in situation 21, SVM and C5.0 in situation 19 had the best performance with the accuracy rate of 95.3%, 92.8% and 90.2%, respectively. In situation 21, both the width and height of segments were 30 pixels. In each part of this situation, there were four segments in each direction. The Table 3, Table 4 and Table 5 for DL, SVM and C5.0 algorithms respectively. Table 6, Table 7 and Table 8 show the confusion matrix for these algorithms when we ignored the confusion between different types of facial expressions. It means that for overall feature number in this situation was 288. In situation 19, we had 30 × 30 pixel segments. The number of segments in each direction was two and we had 72 features. Their confusion matrixes while 12 different classes of facial expression are used are shown in example, both types of happiness type 1 and 2 are happiness and it is not important which type of happiness is recognized. Their overall averages were also in the same rank.

Table 3. Confusion matrix of DL in situation 21

| | AngerT1 | AngerT2 | Disgust | FearT1 | FearT2 | HappinessT1 | HappinessT2 | SadnessT1 | SadnessT2 | SadnessT3 | SadnessT4 | Surprise |
|---|---|---|---|---|---|---|---|---|---|---|---|---|
| AngerT1 * | 81.5 ± 5.7 | 3.4 ± 1.3 | 12.5 ± 2.0 | 0.0 ± 0.0 | 0.0 ± 0.0 | 2.7 ± 0.8 | 0.0 ± 0.0 | 0.0 ± 0.0 | 0.0 ± 0.0 | 0.0 ± 0.0 | 0.7 ± 0.4 | 0.0 ± 0.0 |
| AngerT2 | 4.5 ± 0.3 | 90.4 ± 6.4 | 0.0 ± 0.0 | 0.0 ± 0.0 | 1.9 ± 0.5 | 0.0 ± 0.0 | 0.0 ± 0.0 | 0.0 ± 0.0 | 1.2 ± 0.3 | 0.0 ± 0.0 | 0.0 ± 0.0 | 1.2 ± 0.5 |
| Disgust | 0.0 ± 0.0 | 0.0 ± 0.0 | 96.7 ± 9.6 | 0.0 ± 0.0 | 2.1 ± 0.5 | 0.0 ± 0.0 | 1.2 ± 0.2 | 0.0 ± 0.0 | 0.0 ± 0.0 | 0.0 ± 0.0 | 0.0 ± 0.0 | 0.0 ± 0.0 |
| FearT1 | 0.0 ± 0.0 | 0.0 ± 0.0 | 0.0 ± 0.0 | 89.4 ± 7.0 | 4.2 ± 1.3 | 2.5 ± 0.8 | 2.5 ± 1.0 | 0.0 ± 0.0 | 0.0 ± 0.0 | 0.0 ± 0.0 | 0.0 ± 0.0 | 1.5 ± 0.9 |
| FearT2 | 0.0 ± 0.0 | 0.6 ± 0.2 | 0.6 ± 0.1 | 0.8 ± 0.2 | 91.8 ± 3.6 | 2.1 ± 0.7 | 2.5 ± 0.1 | 0.0 ± 0.0 | 0.0 ± 0.0 | 0.0 ± 0.0 | 0.0 ± 0.0 | 1.6 ± 0.1 |
| HappinessT1 | 0.0 ± 0.0 | 0.0 ± 0.0 | 0.0 ± 0.0 | 0.0 ± 0.0 | 0.0 ± 0.0 | 98.2 ± 4.9 | 1.8 ± 0.2 | 0.0 ± 0.0 | 0.0 ± 0.0 | 0.0 ± 0.0 | 0.0 ± 0.0 | 0.0 ± 0.0 |
| HappinessT2 | 0.0 ± 0.0 | 0.0 ± 0.0 | 0.0 ± 0.0 | 0.6 ± 1.0 | 1.6 ± 0.5 | 4.5 ± 1.3 | 92.4 ± 5.7 | 0.0 ± 0.0 | 0.0 ± 0.0 | 0.0 ± 0.0 | 0.0 ± 0.0 | 0.9 ± 0.0 |
| SadnessT1 | 0.0 ± 0.0 | 0.0 ± 0.0 | 1.0 ± 0.4 | 0.0 ± 0.0 | 0.0 ± 0.0 | 0.9 ± 0.1 | 0.0 ± 0.0 | 82.8 ± 6.4 | 1.3 ± 0.4 | 2.6 ± 0.9 | 9.1 ± 0.9 | 2.3 ± 0.6 |
| SadnessT2 | 0.0 ± 0.0 | 0.0 ± 0.0 | 0.0 ± 0.0 | 0.0 ± 0.0 | 0.0 ± 0.0 | 0.0 ± 0.0 | 0.0 ± 0.0 | 2.6 ± 1.3 | 85.3 ± 6.4 | 1.6 ± 0.8 | 6.2 ± 2.0 | 4.3 ± 1.5 |
| SadnessT3 | 0.0 ± 0.0 | 1.9 ± 1.9 | 0.0 ± 0.0 | 0.0 ± 0.0 | 3.6 ± 2.1 | 0.0 ± 0.0 | 0.0 ± 0.0 | 2.1 ± 1.1 | 2.1 ± 0.5 | 85.6 ± 5.7 | 2.1 ± 0.4 | 2.6 ± 3.3 |
| SadnessT4 | 3.2 ± 0.8 | 1.4 ± 0.4 | 0.0 ± 0.0 | 0.0 ± 0.0 | 0.0 ± 0.0 | 0.0 ± 0.0 | 0.0 ± 0.0 | 2.6 ± 0.9 | 10.5 ± 1.5 | 5.3 ± 0.5 | 75.2 ± 4.9 | 1.8 ± 0.3 |
| Surprise | 0.0 ± 0.0 | 0.0 ± 0.0 | 0.0 ± 0.0 | 0.7 ± 0.1 | 0.0 ± 0.0 | 0.0 ± 0.0 | 0.0 ± 0.0 | 0.0 ± 0.0 | 0.1 ± 0.1 | 0.1 ± 0.0 | 0.0 ± 0.0 | 99.1 ± 7.3 |

* Character T is the abbreviation of "Type"

Table 4. Confusion matrix of SVM in situation 19

| | AngerT1 | AngerT2 | Disgust | FearT1 | FearT2 | HappinessT1 | HappinessT2 | SadnessT1 | SadnessT2 | SadnessT3 | SadnessT4 | Surprise |
|---|---|---|---|---|---|---|---|---|---|---|---|---|
| AngerT1 | 79.2 ± 6.9 | 8.3 ± 3.5 | 3.0 ± 0.5 | 0.0 ± 0.0 | 0.7 ± 0.2 | 0.0 ± 0.0 | 0.0 ± 0.0 | 0.0 ± 0.0 | 1.8 ± 0.5 | 1.8 ± 0.8 | 5.3 ± 1.2 | 0.0 ± 0.0 |
| AngerT2 | 15.0 ± 2.1 | 78.6 ± 4.9 | 0.0 ± 0.0 | 0.0 ± 0.0 | 1.3 ± 0.9 | 0.0 ± 0.0 | 0.0 ± 0.0 | 0.0 ± 0.0 | 0.0 ± 0.0 | 2.5 ± 0.4 | 2.5 ± 0.6 | 0.0 ± 0.0 |
| Disgust | 1.7 ± 0.9 | 0.9 ± 0.5 | 93.9 ± 8.6 | 0.2 ± 0.3 | 0.5 ± 0.4 | 0.0 ± 0.0 | 1.4 ± 3.5 | 0.0 ± 0.0 | 1.8 ± 2.3 | 0.0 ± 0.0 | 0.0 ± 0.0 | 0.0 ± 0.0 |
| FearT1 | 0.0 ± 0.0 | 2.6 ± 2.7 | 0.0 ± 0.0 | 83.4 ± 3.5 | 2.8 ± 0.6 | 0.0 ± 0.0 | 6.5 ± 2.5 | 1.9 ± 0.9 | 0.0 ± 0.0 | 0.0 ± 0.0 | 0.0 ± 0.0 | 2.8 ± 3.1 |
| FearT2 | 4.4 ± 3.2 | 1.2 ± 0.8 | 0.0 ± 0.0 | 5.4 ± 2.7 | 80.5 ± 6.5 | 0.0 ± 0.0 | 7.7 ± 2.4 | 0.0 ± 0.0 | 0.0 ± 0.0 | 0.0 ± 0.0 | 0.9 ± 0.6 | 0.0 ± 0.0 |
| HappinessT1 | 0.0 ± 0.0 | 0.0 ± 0.0 | 1.8 ± 1.4 | 0.0 ± 0.0 | 0.8 ± 0.3 | 91.8 ± 9.8 | 4.5 ± 2.6 | 0.0 ± 0.0 | 0.0 ± 0.0 | 0.0 ± 0.0 | 1.2 ± 0.8 | 0.0 ± 0.0 |
| HappinessT2 | 0.0 ± 0.0 | 0.0 ± 0.0 | 0.0 ± 0.0 | 0.9 ± 0.2 | 3.9 ± 0.7 | 6.8 ± 2.4 | 87.7 ± 4.8 | 0.0 ± 0.0 | 0.0 ± 0.0 | 0.0 ± 0.0 | 0.7 ± 0.5 | 0.0 ± 0.0 |
| SadnessT1 | 0.0 ± 0.0 | 0.0 ± 0.0 | 2.1 ± 1.9 | 0.0 ± 0.0 | 0.0 ± 0.0 | 2.1 ± 2.8 | 0.0 ± 0.0 | 79.1 ± 8.7 | 9.5 ± 3.1 | 2.9 ± 0.7 | 2.4 ± 3.5 | 1.9 ± 0.9 |
| SadnessT2 | 0.0 ± 0.0 | 0.0 ± 0.0 | 2.4 ± 4.5 | 0.0 ± 0.0 | 0.0 ± 0.0 | 2.4 ± 1.8 | 0.0 ± 0.0 | 0.0 ± 0.0 | 75.4 ± 9.1 | 0.0 ± 0.0 | 20.0 ± 5.9 | 0.0 ± 0.0 |
| SadnessT3 | 0.0 ± 0.0 | 5.1 ± 4.8 | 0.0 ± 0.0 | 2.8 ± 3.6 | 0.0 ± 0.0 | 0.0 ± 0.0 | 0.0 ± 0.0 | 6.3 ± 4.5 | 0.0 ± 0.0 | 76.9 ± 8.0 | 6.3 ± 1.5 | 2.7 ± 2.2 |
| SadnessT4 | 3.4 ± 0.9 | 1.1 ± 0.3 | 0.0 ± 0.0 | 0.0 ± 0.0 | 0.0 ± 0.0 | 0.0 ± 0.0 | 0.0 ± 0.0 | 0.0 ± 0.0 | 7.8 ± 6.5 | 0.0 ± 0.0 | 87.8 ± 8.6 | 0.0 ± 0.0 |
| Surprise | 0.0 ± 0.0 | 0.0 ± 0.0 | 0.0 ± 0.0 | 1.2 ± 0.8 | 0.0 ± 0.0 | 0.0 ± 0.0 | 0.6 ± 0.7 | 0.0 ± 0.0 | 0.0 ± 0.0 | 0.6 ± 0.2 | 0.0 ± 0.0 | 97.6 ± 7.4 |
| | 0.0 ± 0.0 | 0.0 ± 0.0 | 0.0 ± 0.0 | 0.0 ± 0.0 | 0.0 ± 0.0 | 0.0 ± 0.0 | 0.0 ± 0.0 | 0.0 ± 0.0 | 0.0 ± 0.0 | 0.0 ± 0.0 | 0.0 ± 0.0 | 0.0 ± 0.0 |

Table 5. Confusion matrix of C5.0 in situation 19

| | AngerT1 | AngerT2 | Disgust | FearT1 | FearT2 | HappinessT1 | HappinessT2 | SadnessT1 | SadnessT2 | SadnessT3 | SadnessT4 | Surprise |
|---|---|---|---|---|---|---|---|---|---|---|---|---|
| AngerT1 | 78.3 ± 7.9 | 12.1 ± 3.5 | 4.4 ± 1.6 | 0.0 ± 0.0 | 0.0 ± 0.0 | 0.0 ± 0.0 | 0.0 ± 0.0 | 0.0 ± 0.0 | 0.0 ± 0.0 | 0.0 ± 0.0 | 5.3 ± 3.5 | 0.0 ± 0.0 |
| AngerT2 | 2.8 ± 3.5 | 86.5 ± 8.7 | 0.0 ± 0.0 | 0.0 ± 0.0 | 2.1 ± 2.9 | 0.0 ± 0.0 | 0.0 ± 0.0 | 0.0 ± 0.0 | 0.0 ± 0.0 | 0.0 ± 0.0 | 8.6 ± 1.2 | 0.0 ± 0.0 |
| Disgust | 1.0 ± 0.2 | 0.2 ± 0.3 | 90.9 ± 4.7 | 0.5 ± 0.7 | 0.2 ± 0.4 | 1.8 ± 0.6 | 1.8 ± 0.7 | 0.0 ± 0.0 | 0.1 ± 0.2 | 0.0 ± 0.0 | 1.1 ± 0.6 | 2.3 ± 0.4 |
| FearT1 | 0.0 ± 0.0 | 1.4 ± 0.9 | 0.0 ± 0.0 | 82.5 ± 5.4 | 4.1 ± 3.6 | 0.0 ± 0.0 | 2.6 ± 2.8 | 0.0 ± 0.0 | 0.0 ± 0.0 | 0.0 ± 0.0 | 0.0 ± 0.0 | 9.2 ± 2.3 |
| FearT2 | 0.0 ± 0.0 | 0.0 ± 0.0 | 0.0 ± 0.0 | 4.9 ± 3.1 | 80.6 ± 7.8 | 1.4 ± 0.9 | 4.8 ± 2.3 | 0.0 ± 0.0 | 0.0 ± 0.0 | 0.0 ± 0.0 | 2.6 ± 4.9 | 6.0 ± 3.7 |
| HappinessT1 | 0.0 ± 0.0 | 0.8 ± 0.4 | 1.4 ± 1.3 | 0.4 ± 0.3 | 0.4 ± 0.6 | 89.6 ± 4.3 | 3.5 ± 1.6 | 0.0 ± 0.0 | 0.6 ± 0.8 | 0.0 ± 0.0 | 2.3 ± 0.9 | 1.0 ± 1.2 |
| HappinessT2 | 0.0 ± 0.0 | 0.0 ± 0.0 | 0.0 ± 0.0 | 1.9 ± 1.6 | 2.9 ± 1.6 | 6.4 ± 4.5 | 85.8 ± 7.1 | 0.0 ± 0.0 | 0.0 ± 0.0 | 0.0 ± 0.0 | 0.0 ± 0.0 | 2.9 ± 1.6 |
| SadnessT1 | 0.0 ± 0.0 | 1.0 ± 1.6 | 8.7 ± 5.3 | 0.6 ± 0.8 | 0.0 ± 0.0 | 10.2 ± 7.4 | 0.0 ± 0.0 | 70.4 ± 3.8 | 0.9 ± 0.7 | 0.0 ± 0.0 | 4.0 ± 2.0 | 4.1 ± 3.1 |
| SadnessT2 | 0.0 ± 0.0 | 0.0 ± 0.0 | 0.0 ± 0.0 | 1.2 ± 2.3 | 0.0 ± 0.0 | 11.9 ± 6.3 | 0.0 ± 0.0 | 0.0 ± 0.0 | 81.9 ± 6.4 | 0.0 ± 0.0 | 4.9 ± 3.6 | 0.1 ± 0.6 |
| SadnessT3 | 1.3 ± 0.9 | 7.2 ± 4.6 | 0.0 ± 0.0 | 0.0 ± 0.0 | 0.9 ± 0.6 | 0.0 ± 0.0 | 0.0 ± 0.0 | 0.0 ± 0.0 | 2.3 ± 2.5 | 81.5 ± 6.8 | 6.9 ± 4.5 | 0.0 ± 0.0 |
| SadnessT4 | 1.0 ± 1.8 | 2.3 ± 3.6 | 8.0 ± 5.3 | 0.0 ± 0.0 | 0.0 ± 0.0 | 8.9 ± 4.6 | 0.0 ± 0.0 | 3.3 ± 2.3 | 1.8 ± 0.9 | 0.0 ± 0.0 | 74.7 ± 9.1 | 0.0 ± 0.0 |
| Surprise | 0.0 ± 0.0 | 0.0 ± 0.0 | 0.0 ± 0.0 | 1.2 ± 2.1 | 0.1 ± 0.9 | 0.0 ± 0.0 | 0.3 ± 0.4 | 0.0 ± 0.0 | 0.0 ± 0.0 | 0.0 ± 0.0 | 0.0 ± 0.0 | 98.4 ± 6.8 |

Table 6. Confusion matrix of DL algorithm in situation 21. Different types of emotions are eliminated.

| | Anger | Disgust | Fear | Happiness | Sadness | Surprise |
|---|---|---|---|---|---|---|
| Anger | 89.9 ± 6.3 | 6.3 ± 2.0 | 1.0 ± 0.1 | 1.3 ± 0.7 | 1.0 ± 0.1 | 0.6 ± 0.5 |
| Disgust | 0.0 ± 0.0 | 96.7 ± 9.6 | 2.1 ± 0.7 | 1.2 ± 0.9 | 0.0 ± 0.0 | 0.0 ± 0.0 |
| Fear | 0.3 ± 0.1 | 0.3 ± 0.1 | 93.1 ± 7.8 | 4.8 ± 0.8 | 0.0 ± 0.0 | 1.5 ± 0.9 |
| Happiness | 0.0 ± 0.0 | 0.0 ± 0.0 | 1.2 ± 0.1 | 98.4 ± 8.6 | 0.0 ± 0.0 | 0.5 ± 0.0 |
| Sadness | 1.6 ± 0.1 | 0.2 ± 0.4 | 0.9 ± 0.1 | 0.2 ± 0.1 | 94.3 ± 5.6 | 2.7 ± 3.7 |
| Surprise | 0.0 ± 0.0 | 0.0 ± 0.0 | 0.7 ± 0.4 | 0.0 ± 0.0 | 0.2 ± 0.1 | 99.1 ± 7.3 |

Table 7. Confusion matrix of SVM algorithm in situation 19. Different types of emotions are eliminated.

| | Anger | Disgust | Fear | Happiness | Sadness | Surprise |
|---|---|---|---|---|---|---|
| Anger | 90.5 ± 6.3 | 1.5 ± 0.5 | 1.0 ± 0.1 | 0.0 ± 0.0 | 7.0 ± 2.6 | 0.0 ± 0.0 |
| Disgust | 2.6 ± 1.1 | 93.9 ± 8.6 | 0.7 ± 0.1 | 1.4 ± 0.6 | 1.4 ± 0.6 | 0.0 ± 0.0 |
| Fear | 4.1 ± 0.9 | 0.0 ± 0.0 | 86.0 ± 3.9 | 7.1 ± 1.7 | 1.4 ± 0.7 | 1.4 ± 3.1 |
| Happiness | 0.0 ± 0.0 | 0.9 ± 1.4 | 2.8 ± 1.2 | 95.4 ± 6.9 | 0.9 ± 0.4 | 0.0 ± 0.0 |
| Sadness | 2.4 ± 0.0 | 1.1 ± 4.9 | 0.7 ± 0.4 | 1.1 ± 0.6 | 93.6 ± 8.4 | 1.1 ± 2.4 |
| Surprise | 0.0 ± 0.0 | 0.0 ± 0.0 | 1.2 ± 0.1 | 0.6 ± 0.3 | 0.6 ± 0.4 | 97.6 ± 7.4 |

Table 8. Confusion matrix of C5.0 algorithm in situation 19. Different types of emotions are eliminated.

| | Anger | Disgust | Fear | Happiness | Sadness | Surprise |
|---|---|---|---|---|---|---|
| Anger | 89.8 ± 8.9 | 2.2 ± 1.6 | 1.0 ± 0.8 | 0.0 ± 0.0 | 6.9 ± 0.8 | 0.0 ± 0.0 |
| Disgust | 1.2 ± 0.6 | 90.9 ± 4.7 | 0.7 ± 1.1 | 3.7 ± 0.6 | 1.2 ± 0.2 | 2.3 ± 0.4 |
| Fear | 0.7 ± 0.1 | 0.0 ± 0.0 | 86.1 ± 8.1 | 4.4 ± 1.3 | 1.3 ± 0.5 | 7.6 ± 4.4 |
| Happiness | 0.4 ± 0.2 | 0.7 ± 1.3 | 2.8 ± 0.2 | 92.7 ± 5.9 | 1.5 ± 0.7 | 1.9 ± 2.0 |
| Sadness | 3.2 ± 0.7 | 4.2 ± 7.5 | 0.7 ± 0.4 | 7.8 ± 1.2 | 83.1 ± 7.2 | 1.0 ± 3.2 |
| Surprise | 0.0 ± 0.0 | 0.0 ± 0.0 | 1.3 ± 0.2 | 0.3 ± 0.4 | 0.0 ± 0.0 | 98.4 ± 6.8 |

As far as misidentifications produced by these methods are concerned, most of them arose from confusion between similar motion vector location and directions. Only the motion vectors direction in area number two and three could distinguish angry Table 3. Other high rate misclassifications in Table 3 happened between fear and happiness. Since both types of happiness and both types of fear had the same motion vectors in the lower part of face, in about five percent, fear was classified Table 3, but also in Table 7 and Table 8. The most important misclassifications are summarized in Table 9 sorted according to the misclassification rate according to the results shown in Table 6, Table 7 and Table 8.

Table 9. The most important misclassifications

| Algorithm name | Emotion | Misclassify as | Rate of misclassification |
|---|---|---|---|
| C5.0 | Sadness | Happiness | 7.8 |
| C5.0 | Fear | Surprise | 7.6 |
| SVM | Fear | Happiness | 7.1 |
| SVM | Anger | Sadness | 7.0 |
| C5.0 | Anger | Sadness | 6.9 |
| DL | Anger | Disgust | 6.3 |
| DL | Fear | Happiness | 4.8 |

According to Table 2, although DL had the best accuracy rate (95.3%) using 288 features, the accuracy of SVM achieves 92.8% using only 72 features. The performance was reduced by only 2.5%. In addition, in situation 19, where DL did the best with the accuracy rate of 92.8%, SVM accuracy was the same as it. In situation 17, where there were only 18 features, SVM worked better than DL. Figure 9 shows the performance fluctuation of algorithms with respect to the number of features. It is clear that increasing the number of features did not improve classification efficiency and in some cases, it decreased the performance. Thus, what must be considered is which algorithm can classify emotion vectors with higher accuracy and lower features.

type 1 from disgust, causing misidentification of them in the case where the motion vector directions were not recognized precisely in these areas. This misclassification is shown in

as happiness according to the results showed in Table 6. This misclassification not only happened in

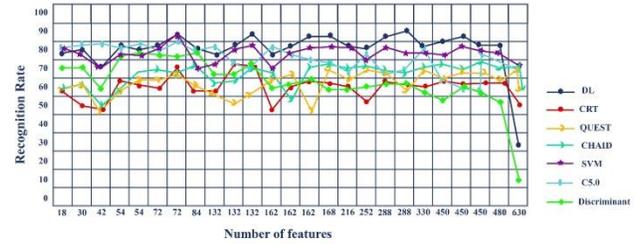

Figure 9. Performance fluctuation of algorithms with respect to the number of features.

Three best situations of each algorithm are shown in Table 10. In three algorithms CRT, SVM and C5.0, situation 19 resulted in the best. It had the second rank in DL. Meanwhile, situation 19 had the best overall accuracy among other situations. As it is clear from Figure 10 (a), in situation 19, the face is divided into four sections in each area. There were two subsections in each direction with equal width and height of 30 pixels. So, it was the best segmentation for facial expression recognition.

Table 10. Three best situations for each algorithm

| DL | CRT | QUEST | CHAID | SVM | C5.0 | Discriminant | Overall AVG |
|---|---|---|---|---|---|---|---|
| 21 | 19 | 16 | 11 | 19 | 19 | 12 | 19 |
| 19 | 20 | 10 | 24 | 17 | 3 | 25 | 18 |
| 23 | 16 | 22 | 6 | 9 | 25 | 15 | 25 |

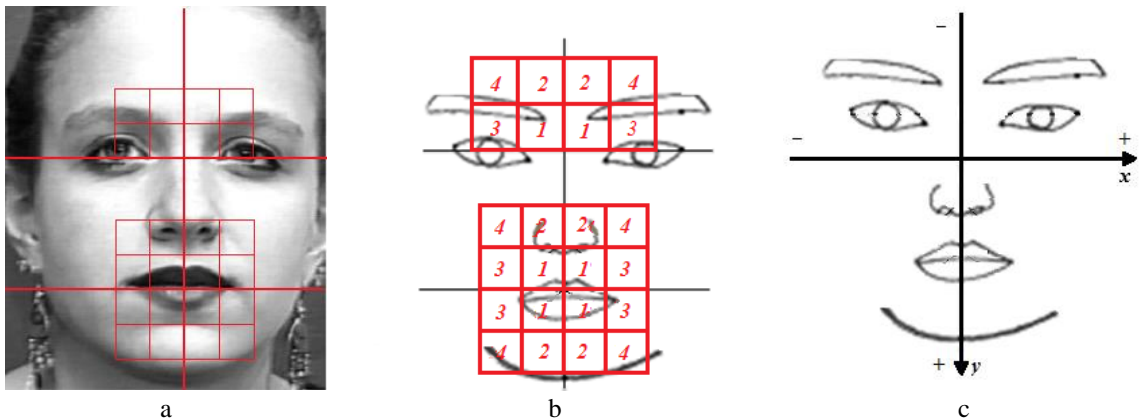

a          b          c

Figure 10. a) Best segmentation to extract features for automatic facial expression recognition b) The number of each segments. c) coordinate system of motion vectors.

If face is segmented as shown in Figure 10 (a) and is numbered as shown in Figure 10 (b), the values of features for different types of facial expression are as in
Table 11. Coordinate system is shown in Figure 10 (c).

Table 11. Values of features in situation 19

| Area number | Segment number | Features | AngerT1 | AngerT2 | Disgust | FearT1 | FearT2 | HappinessT1 | HappinessT2 | SadnessT1 | SadnessT2 | SadnessT3 | SadnessT4 | Surprise |
|---|---|---|---|---|---|---|---|---|---|---|---|---|---|---|
| 1 | 1 | P | 0.02 | 0.02 | 0.02 | 0.02 | 0.02 | 0.02 | 0.01 | 0.02 | 0.02 | 0.02 | 0.02 | 0.02 |
| | | LX | 0.13 | 0.08 | 0.01 | -0.01 | 0.06 | 0 | -0.01 | 0.09 | -0.01 | -0.08 | 0.06 | 0.01 |
| | | LY | 0.43 | 0.45 | 0.61 | -0.59 | 0.16 | 0.01 | 0.03 | -0.37 | 0.02 | -0.17 | 0.1 | -1.21 |
| | 2 | P | 0.02 | 0.02 | 0.02 | 0.02 | 0.02 | 0.01 | 0.01 | 0.02 | 0.01 | 0.02 | 0.02 | 0.02 |
| | | LX | 0.14 | 0.15 | 0.07 | -0.05 | 0.06 | 0 | 0.01 | 0.05 | 0 | -0.09 | 0.06 | -0.05 |
| | | LY | 0.48 | 0.6 | 0.8 | -0.83 | 0.19 | 0.01 | 0 | -0.57 | 0.02 | -0.35 | 0.13 | -1.55 |
| | 3 | P | 0.02 | 0.02 | 0.02 | 0.02 | 0.02 | 0.02 | 0.02 | 0.02 | 0.02 | 0.03 | 0.02 | 0.02 |
| | | LX | 0.27 | 0.16 | 0.07 | 0 | 0.08 | 0 | -0.01 | 0.08 | 0 | -0.06 | 0.1 | 0.01 |
| | | LY | 0.51 | 0.47 | 0.51 | -0.42 | 0.27 | -0.01 | 0.05 | -0.19 | 0 | -0.05 | 0.27 | -1.15 |
| | 4 | P | 0.02 | 0.02 | 0.02 | 0.02 | 0.02 | 0.01 | 0.01 | 0.02 | 0.01 | 0.02 | 0.02 | 0.02 |
| | | LX | 0.43 | 0.43 | 0.17 | -0.12 | 0.22 | -0.01 | -0.02 | 0.1 | -0.02 | -0.05 | 0.16 | -0.11 |
| | | LY | 0.64 | 0.72 | 0.69 | -0.9 | 0.36 | 0.02 | 0.04 | -0.43 | 0.01 | -0.19 | 0.25 | -1.75 |
| 2 | 1 | P | 0.02 | 0.03 | 0.02 | 0.02 | 0.02 | 0.02 | 0.02 | 0.02 | 0.02 | 0.02 | 0.02 | 0.02 |
| | | LX | -0.12 | -0.17 | -0.02 | -0.06 | -0.13 | 0 | 0 | -0.09 | -0.02 | -0.15 | -0.09 | -0.04 |
| | | LY | 0.4 | 0.46 | 0.59 | -0.52 | 0.18 | 0.01 | 0.02 | -0.32 | 0.02 | -0.09 | 0.14 | -1.2 |
| | 2 | P | 0.02 | 0.02 | 0.02 | 0.02 | 0.02 | 0.01 | 0.01 | 0.02 | 0.01 | 0.02 | 0.02 | 0.02 |
| | | LX | -0.15 | -0.22 | -0.07 | -0.06 | -0.18 | 0.01 | 0 | -0.03 | -0.01 | -0.18 | -0.11 | 0.03 |
| | | LY | 0.45 | 0.63 | 0.78 | -0.83 | 0.18 | 0.02 | 0.04 | -0.62 | 0.01 | -0.34 | 0.12 | -1.6 |
| | 3 | P | 0.02 | 0.02 | 0.02 | 0.02 | 0.02 | 0.02 | 0.02 | 0.02 | 0.02 | 0.03 | 0.02 | 0.02 |
| | | $LX_{20}$ | -0.17 | -0.21 | -0.07 | -0.06 | -0.12 | 0.01 | 0 | -0.08 | -0.01 | -0.14 | -0.13 | 0.02 |
| | | LY | 0.45 | 0.5 | 0.45 | -0.35 | 0.27 | 0 | 0.04 | -0.22 | 0.01 | 0.1 | 0.26 | -1.07 |
| | 4 | P | 0.02 | 0.02 | 0.02 | 0.02 | 0.02 | 0.02 | 0.01 | 0.02 | 0.01 | 0.02 | 0.02 | 0.02 |
| | | LX | -0.35 | -0.41 | -0.13 | -0.01 | -0.26 | 0.03 | 0 | -0.05 | -0.01 | -0.2 | -0.18 | 0.07 |
| | | LY | 0.66 | 0.75 | 0.66 | -0.78 | 0.41 | 0.04 | 0.05 | -0.56 | 0.03 | -0.12 | 0.23 | -1.77 |
| 3 | 1 | P | 0.02 | 0.02 | 0.02 | 0.02 | 0.02 | 0.02 | 0.02 | 0.02 | 0.02 | 0.03 | 0.02 | 0.02 |
| | | LX | 0.09 | -0.02 | -0.06 | -0.14 | -0.22 | -0.18 | -0.2 | 0 | -0.01 | 0.13 | 0 | -0.04 |
| | | LY | -0.04 | 0.19 | -0.84 | -0.08 | -0.02 | -0.5 | -0.61 | -0.1 | -0.07 | 0.5 | 0.09 | -0.48 |
| | 2 | P | 0.02 | 0.02 | 0.02 | 0.02 | 0.02 | 0.02 | 0.02 | 0.02 | 0.02 | 0.03 | 0.02 | 0.02 |
| | | LX | 0.01 | -0.04 | -0.07 | -0.1 | -0.13 | -0.1 | -0.18 | 0.03 | -0.01 | 0.05 | 0 | -0.02 |
| | | LY | 0.03 | 0.19 | -0.56 | -0.23 | 0.06 | -0.11 | -0.2 | 0.05 | 0.17 | 0.3 | 0.27 | -0.78 |
| | 3 | P | 0.02 | 0.02 | 0.02 | 0.02 | 0.02 | 0.02 | 0.02 | 0.02 | 0.02 | 0.02 | 0.02 | 0.02 |
| | | LX | 0.04 | -0.02 | 0.09 | -0.44 | -0.44 | -0.59 | -0.76 | 0 | -0.08 | 0.19 | -0.04 | 0.23 |
| | | LY | -0.15 | 0.12 | -0.92 | -0.11 | -0.12 | -1 | -1.09 | 0.13 | 0.35 | 0.62 | 0.35 | 0.16 |
| | 4 | P | 0.02 | 0.02 | 0.02 | 0.02 | 0.02 | 0.02 | 0.02 | 0.02 | 0.02 | 0.02 | 0.02 | 0.02 |
| | | LX | 0.03 | 0.01 | -0.01 | -0.23 | -0.21 | -0.32 | -0.43 | 0.09 | 0.02 | 0.11 | 0.04 | 0.04 |
| | | LY | -0.02 | 0.11 | -0.98 | -0.27 | -0.15 | -0.55 | -0.72 | 0.1 | 0.24 | 0.32 | 0.27 | -0.37 |
| 4 | 1 | P | 0.02 | 0.02 | 0.02 | 0.02 | 0.02 | 0.02 | 0.02 | 0.02 | 0.02 | 0.03 | 0.02 | 0.02 |
| | | LX | -0.01 | -0.04 | 0.02 | 0.25 | 0.24 | 0.27 | 0.35 | -0.01 | 0.03 | -0.09 | 0.03 | -0.03 |
| | | LY | -0.05 | 0.16 | -0.87 | -0.11 | -0.05 | -0.5 | -0.61 | -0.06 | -0.06 | 0.5 | 0.09 | -0.46 |
| | 2 | P | 0.02 | 0.03 | 0.02 | 0.02 | 0.02 | 0.02 | 0.02 | 0.02 | 0.02 | 0.03 | 0.02 | 0.02 |
| | | LX | 0 | -0.03 | 0.04 | 0.11 | 0.12 | 0.15 | 0.21 | -0.02 | -0.01 | -0.08 | -0.03 | -0.02 |
| | | LY | 0.03 | 0.17 | -0.62 | -0.24 | 0.05 | -0.15 | -0.22 | 0.08 | 0.16 | 0.27 | 0.27 | -0.76 |

|   |   |    | 1 | 2 | 3 | 4 | 5 | 6 | 7 | 8 | 9 | 10 | 11 | 12 |
|---|---|----|---|---|---|---|---|---|---|---|---|----|----|----|
|   |   | P  | 0.02 | 0.02 | 0.02 | 0.02 | 0.02 | 0.02 | 0.02 | 0.02 | 0.02 | 0.02 | 0.02 | 0.02 |
|   | 3 | LX | 0.02 | -0.04 | -0.14 | 0.44 | 0.43 | 0.58 | 0.8 | 0.02 | 0.02 | -0.23 | 0.06 | -0.28 |
|   |   | LY | -0.2 | 0.16 | -0.93 | -0.14 | -0.17 | -1 | -1.07 | 0.26 | 0.33 | 0.68 | 0.4 | 0.26 |
|   |   | P  | 0.02 | 0.02 | 0.02 | 0.02 | 0.02 | 0.02 | 0.02 | 0.02 | 0.02 | 0.02 | 0.02 | 0.02 |
|   | 4 | LX | 0.04 | -0.05 | -0.08 | 0.17 | 0.19 | 0.34 | 0.45 | -0.07 | -0.09 | -0.21 | -0.03 | -0.09 |
|   |   | LY | -0.04 | 0.13 | -0.98 | -0.29 | -0.14 | -0.57 | -0.7 | 0.14 | 0.25 | 0.4 | 0.27 | -0.31 |
| 5 |   | P  | 0.02 | 0.02 | 0.02 | 0.02 | 0.02 | 0.02 | 0.02 | 0.02 | 0.02 | 0.02 | 0.02 | 0.01 |
|   | 1 | LX | 0.16 | -0.02 | -0.02 | -0.21 | -0.26 | -0.14 | -0.23 | -0.15 | -0.04 | -0.05 | -0.05 | -0.02 |
|   |   | LY | -0.57 | 0.12 | -0.64 | 1 | 0.86 | -0.45 | 0.65 | -0.5 | -0.56 | 0.39 | -0.37 | 1.31 |
|   |   | P  | 0.01 | 0.01 | 0.01 | 0.01 | 0.01 | 0.02 | 0.02 | 0.02 | 0.02 | 0.01 | 0.02 | 0.01 |
|   | 2 | LX | 0.09 | -0.06 | 0.05 | -0.19 | -0.26 | -0.12 | -0.23 | 0.09 | 0.08 | 0.05 | 0.04 | -0.03 |
|   |   | LY | -0.65 | 0.1 | -0.47 | 1.05 | 0.97 | -0.33 | 0.87 | -0.95 | -1.09 | -0.06 | -0.87 | 2.07 |
|   |   | P  | 0.01 | 0.02 | 0.02 | 0.02 | 0.02 | 0.02 | 0.02 | 0.02 | 0.02 | 0.02 | 0.02 | 0.02 |
|   | 3 | LX | 0.17 | -0.04 | 0.08 | -0.69 | -0.71 | -0.42 | -0.73 | -0.09 | -0.16 | 0.06 | -0.11 | 0.34 |
|   |   | LY | -0.38 | 0.1 | -0.64 | 0.43 | 0.28 | -1 | -0.5 | 0.02 | 0.23 | 0.64 | 0.21 | 1.09 |
|   |   | P  | 0.01 | 0.01 | 0.01 | 0.01 | 0.01 | 0.01 | 0.01 | 0.01 | 0.02 | 0.01 | 0.01 | 0.01 |
|   | 4 | LX | 0.07 | -0.04 | 0.07 | -0.58 | -0.56 | -0.18 | -0.51 | 0.13 | 0.11 | 0.06 | 0.11 | 0.03 |
|   |   | LY | -0.47 | 0.01 | -0.34 | 0.57 | 0.53 | -0.59 | 0.17 | -0.39 | -0.4 | 0.34 | -0.3 | 1.5 |
| 6 |   | P  | 0.02 | 0.02 | 0.02 | 0.02 | 0.02 | 0.02 | 0.02 | 0.02 | 0.02 | 0.03 | 0.02 | 0.01 |
|   | 1 | LX | -0.12 | -0.09 | -0.08 | 0.35 | 0.32 | 0.18 | 0.36 | -0.11 | 0.04 | -0.11 | 0.06 | 0.03 |
|   |   | LY | -0.61 | 0.15 | -0.64 | 0.91 | 0.85 | -0.42 | 0.55 | -0.48 | -0.44 | 0.4 | -0.28 | 1.32 |
|   |   | P  | 0.01 | 0.01 | 0.01 | 0.02 | 0.01 | 0.02 | 0.02 | 0.02 | 0.02 | 0.02 | 0.02 | 0.01 |
|   | 2 | LX | -0.02 | -0.12 | -0.05 | 0.36 | 0.33 | 0.12 | 0.33 | -0.2 | -0.18 | -0.1 | -0.1 | 0.08 |
|   |   | LY | -0.67 | 0.08 | -0.5 | 1.01 | 0.97 | -0.29 | 0.83 | -1 | -1 | -0.04 | -0.81 | 2.04 |
|   |   | P  | 0.02 | 0.02 | 0.02 | 0.02 | 0.02 | 0.02 | 0.02 | 0.02 | 0.02 | 0.02 | 0.02 | 0.02 |
|   | 3 | LX | -0.14 | -0.06 | -0.13 | 0.76 | 0.71 | 0.39 | 0.79 | 0.01 | 0.07 | -0.12 | 0.1 | -0.28 |
|   |   | LY | -0.4 | 0.14 | -0.65 | 0.35 | 0.24 | -0.94 | -0.53 | 0.04 | 0.11 | 0.6 | 0.26 | 1.17 |
|   |   | P  | 0.01 | 0.01 | 0.01 | 0.01 | 0.01 | 0.01 | 0.01 | 0.02 | 0.02 | 0.01 | 0.01 | 0.01 |
|   | 4 | LX | -0.13 | -0.04 | -0.06 | 0.52 | 0.53 | 0.16 | 0.49 | -0.25 | -0.16 | -0.13 | -0.02 | 0.06 |
|   |   | LY | -0.43 | 0.03 | -0.39 | 0.44 | 0.48 | -0.56 | 0 | -0.44 | -0.29 | 0.23 | -0.2 | 1.49 |

Extracted features were ranked according to their importance in the classification shown in To show how these features retain the emotion, in Figure 11, a low (three) dimension space with PCA Table 12. Information gain [79] was used for ranking features. There are some amazing points in this table. For example, among the first top 10 features, almost all of them were motion vector length in Y direction. Among the top 50 features, 48% were motion vector length in Y direction, 42% were motion vector length in X direction and only 10% represented the ratio of motion vector number in different areas. It was clear that length in Y directions was the most distinctive feature. Another amazing point about this table is that area #3 and area #4 in Figure 4(a) had fewer role than other areas in the classification. To show how these features retain the emotion, in Figure 11, a low (three) dimension space with PCA [80], PLS [81] and t-SNE [82] are illustrated within different views.

Table 12. Rank of features according to their importance in classification. Feature indexes i and j are the area number and the section number, respectively. For example, $LY_{14}$ means motion vector length in Y direction in area 1, section 4.

| Rank | Feature | Rank | Feature | Rank | Feature | Rank | Feature | Rank | Feature | Rank | Feature |
|------|---------|------|---------|------|---------|------|---------|------|---------|------|---------|
| 1 | $LY_{14}$ | 11 | $LY_{54}$ | 21 | $LY_{61}$ | 31 | $LX_{14}$ | 41 | $LX_{51}$ | 51 | $LX_{12}$ |
| 2 | $LY_{62}$ | 12 | $LY_{64}$ | 22 | $LY_{33}$ | 32 | $LX_{62}$ | 42 | $P_{51}$ | 52 | $P_{41}$ |
| 3 | $LY_{24}$ | 13 | $LY_{63}$ | 23 | $LX_{44}$ | 33 | $LX_{61}$ | 43 | $P_{61}$ | 53 | $LX_{21}$ |
| 4 | $LY_{22}$ | 14 | $LX_{53}$ | 24 | $LX_{54}$ | 34 | $LX_{41}$ | 44 | $P_{12}$ | 54 | $P_{31}$ |
| 5 | $LY_{52}$ | 15 | $LX_{33}$ | 25 | $LX_{34}$ | 35 | $LY_{31}$ | 45 | $LX_{13}$ | 55 | $P_{54}$ |
| 6 | $LY_{12}$ | 16 | $LY_{53}$ | 26 | $LY_{44}$ | 36 | $LY_{41}$ | 46 | $LX_{32}$ | 56 | $P_{24}$ |
| 7 | $LY_{21}$ | 17 | $LY_{23}$ | 27 | $LY_{34}$ | 37 | $LX_{24}$ | 47 | $P_{62}$ | 57 | $P_{22}$ |
| 8 | $LY_{11}$ | 18 | $LY_{51}$ | 28 | $LY_{32}$ | 38 | $LX_{42}$ | 48 | $LX_{22}$ | 58 | $P_{14}$ |
| 9 | $LY_{13}$ | 19 | $LX_{63}$ | 29 | $LX_{64}$ | 39 | $LX_{52}$ | 49 | $LX_{23}$ | 59 | $LX_{11}$ |
| 10 | $LX_{43}$ | 20 | $LY_{43}$ | 30 | $LY_{42}$ | 40 | $LX_{31}$ | 50 | $P_{52}$ | 60 | $P_{33}$ |

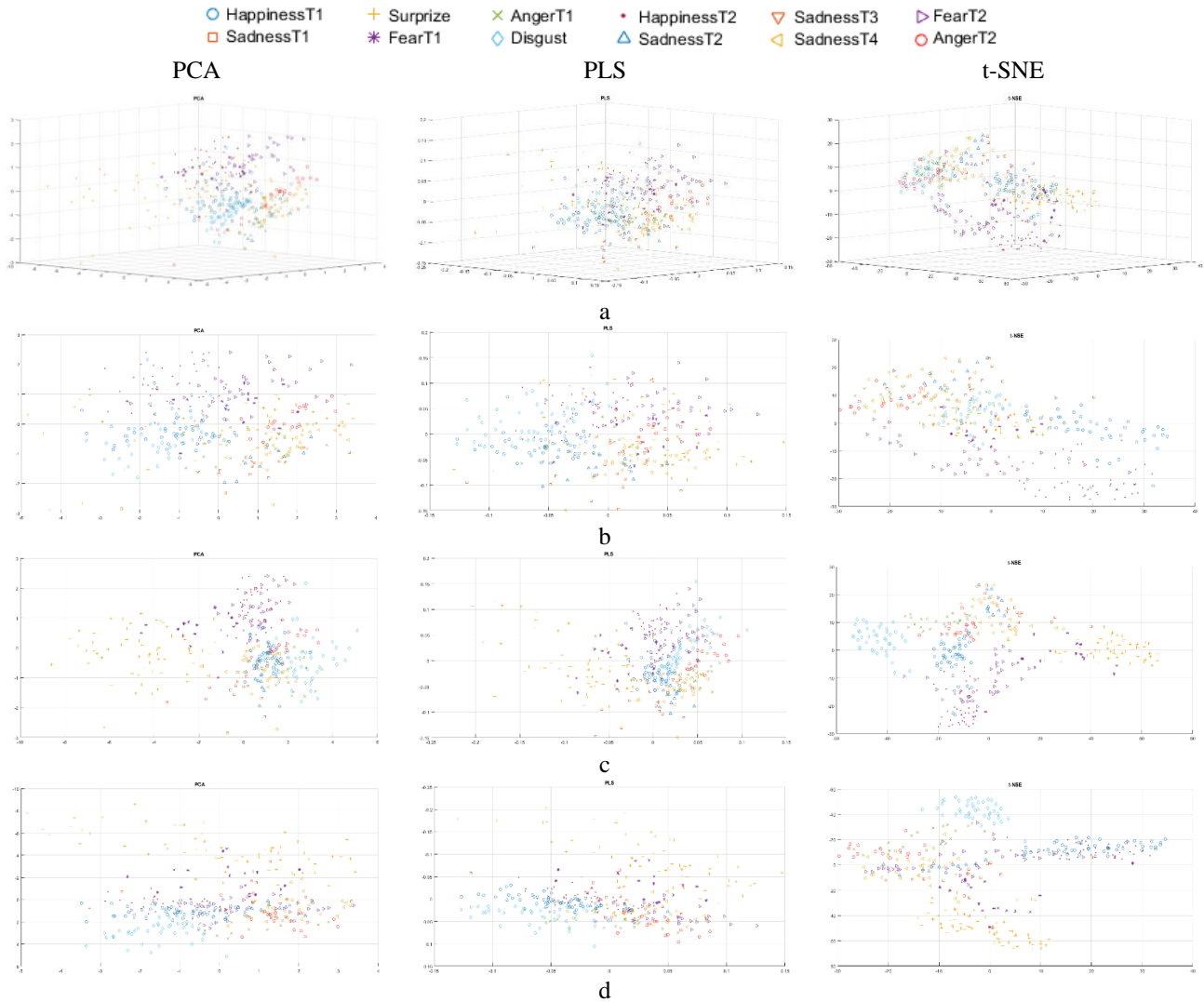

Figure 11. Three dimension space with PCA, PLS and t-NSE within different views while azimuth and elevation are a) 45 and 10 b) 90 and 0 c) 0 and 0 d) 90 and 90, respectively.

## VI. CONCLUSIONS AND FUTURE WORK

Automating the analysis of facial changes, especially from a frontal view, is important to advance the studies on automatic facial expression recognition, design human-machine interfaces, and boost some applications in areas such as security, medicine, making animations and education. In this research, some of the most famous classification algorithms were used upon changes in the position of facial points. These points were tracked in an image sequence of a frontal view of the face. The best methods were chosen. They were DL, SVM and C5.0, with the accuracy rate of 95.3%, 92.8% and 90.2%, respectively. The most distinguishing changes were found to be deformation in Y direction, in the upper and lower areas of the face. Meanwhile, some more changes of face during facial expression were investigated in this research. It shows that six more changes can be identified in addition to six basic changes happens in face during representation of facial expression. This paper not only provided a basic understanding of how facial points could change during a facial expression, but also it tried to classify these deformations.

Future work on this issue aims at investigating automatic facial expression while head has rigid motions. Meanwhile, the performance of the method must be invariant to occlusions like glasses and facial hair. In addition, the method must perform well independently of the changes in the illumination intensity while image sequence is created.